\def\year{2021}\relax
\documentclass[letterpaper]{article} 
\usepackage{aaai21}  
\usepackage{times}  
\usepackage{helvet} 
\usepackage{courier}  
\usepackage[hyphens]{url}  
\usepackage{graphicx} 
\usepackage{natbib}  
\usepackage{caption} 
\usepackage{url}
\usepackage{amsfonts}
\usepackage{microtype}      
\usepackage[noend]{algpseudocode}
\usepackage{algorithmicx,algorithm}
\usepackage{amsmath}
\usepackage{epstopdf}
\usepackage{subfigure}
\usepackage{booktabs}
\usepackage{bm,bbm}
\usepackage{color}
\usepackage{wrapfig}
\usepackage{array}
\usepackage{hyperref}

\usepackage{lipsum} 
\newcommand{\etal}{\textit{et al}.}
\newcommand{\ie}{\textit{i}.\textit{e}.,}
\newcommand{\eg}{\textit{e}.\textit{g}.,}
\nocopyright

\title{Diversity Helps: Unsupervised Few-shot Learning via Distribution Shift-based Data Augmentation}
\author{Tiexin Qin\thanks{contributed equally},\quad Wenbin Li\footnotemark[1],\quad Yinghuan Shi,\quad Yang Gao \\
State Key Laboratory for Novel Software Technology, Nanjing University, China\\
\tt\small qtx@smail.nju.edu.cn \quad \tt\small \{liwenbin,syh,gaoy\}@nju.edu.cn}

\begin{document}
\maketitle
\begin{abstract}
Few-shot learning aims to learn a new concept when only a few training examples are available, which has been extensively explored in recent years. However, most of the current works heavily rely on a large-scale labeled auxiliary set to train their models in an episodic-training paradigm. Such a kind of supervised setting basically limits the widespread use of few-shot learning algorithms.
Instead, in this paper, we develop a novel framework called \emph{Unsupervised Few-shot Learning via Distribution Shift-based Data Augmentation} (ULDA), which pays attention to the distribution diversity inside each constructed pretext few-shot task when using data augmentation. Importantly, we highlight the value and importance of the distribution diversity in the augmentation-based pretext few-shot tasks, which can effectively alleviate the overfitting problem and make the few-shot model learn more robust feature representations.
In ULDA, we systemically investigate the effects of different augmentation techniques and propose to strengthen the distribution diversity (or difference) between the query set and support set in each few-shot task, by augmenting these two sets diversely (\ie~distribution shifting). In this way, even incorporated with simple augmentation techniques (\eg~random crop, color jittering, or rotation), our ULDA can produce a significant improvement. In the experiments, few-shot models learned by ULDA can achieve superior generalization performance and obtain state-of-the-art results in a variety of established few-shot learning tasks on Omniglot and \emph{mini}ImageNet. The source code is available in \textcolor{blue}{\emph{https://github.com/WonderSeven/ULDA}}.
\end{abstract}

\section{Introduction}

The ability of learning from limited labeled examples is a hallmark of human intelligence, yet it remains a challenge for modern machine learning systems. This problem recently has attracted significant attention from the machine learning community, which is formalized as few-shot learning (FSL). To solve this problem, a large-scale auxiliary set is generally required to learn transferable knowledge to boost the learning of the target few-shot tasks. Specifically, one kind of FSL methods usually resort to using metric losses to enhance the discriminability of the representation learning, such that a simple nearest neighbor or linear classifier is able to achieve satisfactory classification results~\cite{Snell2016NIPS,Vinyals2016NIPS}. Another kind of FSL methods incorporates the concept of meta-learning and aims to enhance the ability of quickly updating with a few labeled examples~\cite{FinnICML2017,RaviICLR2017,MunkhdalaiICML2017}. Alternatively, some FSL methods address this problem by generating more examples from the provided ones~\cite{Gao2018NIPS,Chen2019AAAI,ChenCVPR2019}.

\begin{table}[!tbp]\small
\normalsize
\vspace*{-1cm}
\begin{center}
\caption{The results of $N$-way $K$-shot tasks on \emph{mini}ImageNet by using different augmentation methods on \emph{ProtoNet} to construct the query and support sets. TA and AA indicate traditional augmentation and AutoAugment, respectively.}
\label{tab:compare_augmentation}
\begin{tabular}{cccc}
\toprule[1pt]
\textbf{Support} &  \textbf{Query} & \textbf{5-way 1-shot} & \textbf{5-way, 5-shot} \\  
\midrule
TA              & TA              & 32.58 & 44.40 \\ 
AA              & AA           & 31.53 & 41.83 \\ 
\textbf{TA}     & \textbf{AA}  & \underline{34.07} & \underline{47.31} \\ 
\textbf{AA}     & \textbf{TA}     & \textbf{35.37} & \textbf{49.16} \\ 
\bottomrule[1pt]
\end{tabular}
\end{center}
\vspace{-0.2cm}
\end{table}

\begin{figure}[tbp]
\begin{center}
\includegraphics[width=0.47\textwidth]{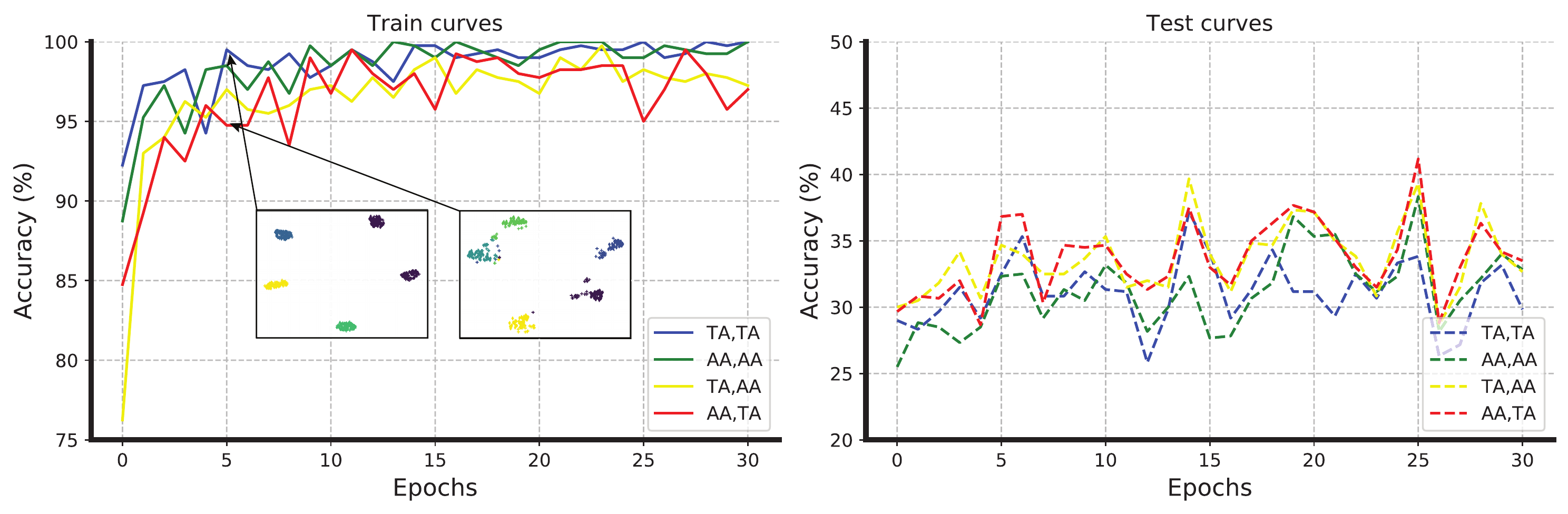}
\end{center}
\vspace{-0.3cm}
\caption{The train and test accuracy curves on the 5-way 1-shot tasks, corresponding to the four combinations of different augmentation methods (\ie~TA and AA) in Table~\ref{tab:compare_augmentation}. As seen, the diverse combinations (\ie~the red and yellow lines) enjoy a smaller risk of overfitting (\ie~lower train accuracy and higher test accuracy) than the identical combinations (\ie~the green and blue lines).}
\label{fig:train_test_curves}
\vspace{-0.3cm}
\end{figure}

Although the aforementioned FSL methods can achieve promising results, most of these methods are fully supervised, which means that they are heavily relying on a large-scale fully labeled auxiliary set (\eg~a subset from ImageNet in previous works~\cite{Snell2016NIPS,FinnICML2017,RaviICLR2017}).
Through this fully labeled auxiliary set, plenty of supervised few-shot tasks (episodes) can be constructed for model training (\ie~episodic-training mechanism~\cite{Vinyals2016NIPS}). However, in many real-world applications, such a fully supervised condition is relatively severe. It greatly hinder the widespread use of these FSL methods for real applications. Because data labeling for a large-scale dataset is normally time-consuming, laborious, and even very expensive for some domain-professional areas like biomedical data analysis. In contrast, large unlabeled data is easily accessible to many real problems. This gives rise to a more challenging problem, called \textit{unsupervised few-shot learning}, which tries to learn few-shot models by using an unlabeled auxiliary set.

As for unsupervised few-shot learning, only a few works have been proposed. For example, CACTUs~\cite{Hsu2019ICLR}, a two-stage method, firstly uses a clustering algorithm to obtain pseudo labels, and then trains a model under the common supervised few-shot setting with these pseudo labels. Different from CACTUs, both AAL~\cite{AAL2019ICML} and UMTRA~\cite{UMTRA2019NIPS} take each instance as one class and randomly sample multiple examples to construct a support set. Next, they generate a pseudo query set according to the support set by leveraging data augmentation techniques. In this paper, we are more interested in this data augmentation based direction, because it can not only achieve promising results but also can be easily learned in an end-to-end manner. However, we find that the existing data augmentation based methods (\ie~AAL and UMTRA) are sensitive to the selection of augmentation techniques and usually do not contain sufficient regularity for model learning. What's more, they are easily suffering from the overfitting problem during training, because they choose the same data augmentation technique for both the query set and support set. This will make the distributions between the augmented query set and support set too similar. In other words, they construct too many ``easy pretext few-shot tasks'' for the downstream few-shot training by only using one single data augmentation technique. 
In some cases, this technique equals to directly copy original samples several times.
We argue that such \textit{excessive distribution similarity} between the query and support set (\ie~easy pretext few-shot tasks) is the main point of leading to overfitting issue in unsupervised few-shot model training.

To tackle the above overfitting problem, we claim that strengthening the \textit{distribution diversity (or difference)} between the augmented query set and support set (\ie~hard pretext few-shot tasks) can significantly alleviate the overfitting problem during the model training and make the learned model have a much better generalization ability. To simply verify this point, we perform a preliminary experiment (see 
Table~\ref{tab:compare_augmentation} and Figure~\ref{fig:train_test_curves}). 
We observe that there is a high risk of overfitting when using same augmentation technique.
Also, when using different (or diverse) augmentation techniques, the classification performance can be significantly improved over using same augmentation technique.



Therefore, in this paper, we introduce a novel framework named \textit{Unsupervised Few-shot Learning via Distribution Shift-based Data Augmentation} (ULDA) following the above statement. To be specific, our ULDA augments the query set and support set in diverse ways, aiming to make a significant distribution shift between these two sets. The main contributions of our work could be summarized into the following three folds: 
\begin{enumerate}
    \item We argue that the \textit{distribution diversity} between the augmented query set and support set is a key point in data augmentation based unsupervised few-shot learning, for the first time in the literature.
    \item  We propose a \textit{Unsupervised Few-shot Learning via Distribution Shift-based Data Augmentation} (ULDA) framework and a new simple augmentation method named \emph{Distribution Shift-based Task Internal Mixing} (DSTIM) to strengthen the distribution diversity when constructing the pretext few-shot training tasks.
    \item Extensive experiments on both Omniglot and \emph{mini}ImageNet datasets demonstrate the superiority of our proposed ULDA and DSTIM.
\end{enumerate}

\section{Related Work}

\label{sec:related_work}
We briefly review the related work about supervised and unsupervised few-shot learning, respectively.

\textbf{Few-shot learning (FSL).} FSL aims to learn a new concept on very limited training examples, which has promising practical application value. A vast number of methods has been proposed in recent years. These methods can be roughly categorized into three classes, \ie~\emph{metric-based}, \emph{optimization-based}, and \emph{hallucination-based methods}.

The metric-based methods aim to learn discriminative feature representations by using deep metric learning, with the help of intra-class and inter-class constraints~\cite{Vinyals2016NIPS,Snell2016NIPS,Sung2018CVPR,Li2019DN4}. They employ various metric losses (\eg~pairwise loss, triplet loss) to enhance the discriminability of the learned features. The optimization-based methods strive for enhancing the flexibility of the learned model such that it can be readily updated with a few labeled examples \cite{RaviICLR2017,FinnICML2017,LeeCVPR2019meta,ChenICLR2019close}. Alternatively, the hallucination-based methods attempt to address the data scarcity problem by directly generating more new examples \cite{ZhangCVPR2019,AlfassyCVPR2019,ChenCVPR2019,Chen2019AAAI,ChenTIP2019}.

Most methods train their models under the episodic-training paradigm \cite{Vinyals2016NIPS}. They organize a large labeled auxiliary dataset into plenty of mimetic few-shot tasks where each task contains a \emph{support} set and a \emph{query} set. The \emph{support} set is used to acquire task-specific information and the \emph{query} set is used to evaluate the generalization performance of the model. Based on episodic-training, the model expects to learn transferable representations or knowledge, with which, it can generalize to new unseen tasks.

\begin{figure*}[!t]
\vspace{-0.2cm}
\centering
\includegraphics[width=0.75\textwidth]{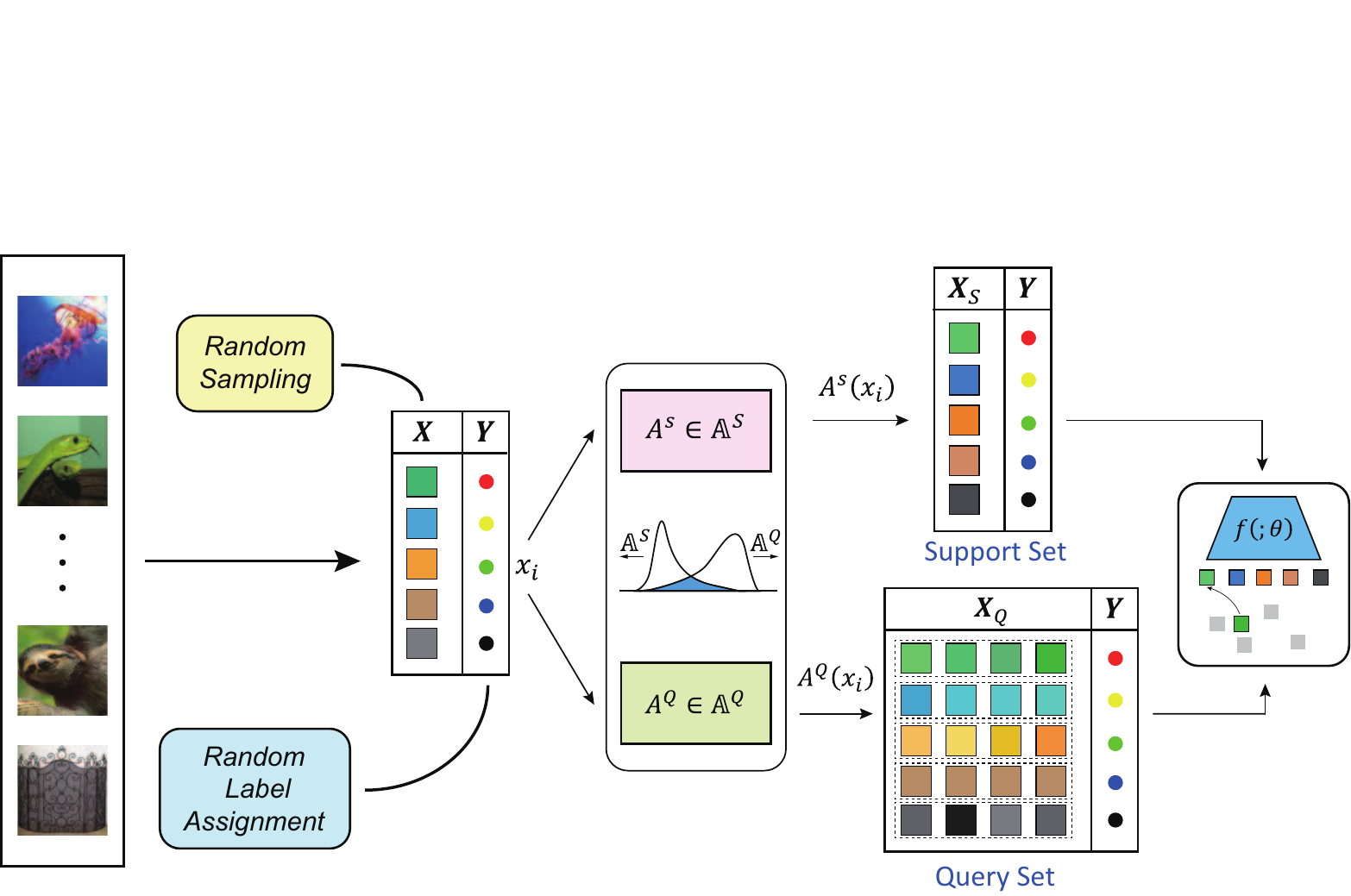}
\caption{The framework of the proposed ULDA which starts from an unlabeled auxiliary dataset. First, randomly select $N$ examples and assign $N$ random labels to them. After that, the proposed \textit{distribution shift-based augmentation} module is used to construct a pretext few-shot task (consists of an augmented query set and an augmented support set). Specifically, the query set and support set are augmented by the augmentation operators $A^Q\in\mathbb{A}^Q$ and $A^S\in\mathbb{A}^S$, respectively. Finally, the constructed pretext few-shot task is adopted to train the few-shot learning model in a supervised way.}
\label{fig:framework}
\vspace{-0.2cm}
\end{figure*}

\textbf{Unsupervised few-shot learning.}
Currently, a few works propose \emph{unsupervised few-shot learning} to tackle the huge requirement of a large labeled auxiliary set in supervised few-shot learning. Hsu \etal~\cite{Hsu2019ICLR} propose CACTUs which uses a clustering algorithm to obtain pseudo labels and then constructs few-shot tasks with these pseudo labels. Differently, Khodadadeh \etal~\cite{UMTRA2019NIPS} and Antoniou \etal~\cite{AAL2019ICML} both propose to randomly sample multiple examples to construct the support set and generate a pseudo query set via data augmentation based on the support set.

Our work belongs to the data augmentation based methods. The main difference is that the existing methods~\cite{UMTRA2019NIPS,AAL2019ICML} easily suffers from the overfitting problem, while our proposed ULDA can significantly alleviate this problem. This is because there is usually a large distribution similarity between the query set and support set in the existing methods, while our ULDA strengthens a distribution shift between the augmented query set and support set. Note that similar operations seemingly have appeared in other research fields, such as FixMatch~\cite{Sohn2020FixMatch} in semi-supervised learning and SimCLR~\cite{Chen2020SimCLR} in unsupervised representation learning. However, we highlight that we do not need to use weak augmentations to assign a higher degree of confidence for unlabeled samples like FixMatch did, and we draw a different observation (diversity helps) from SimCLR (combination is better). In fact, we believe that our observation/perspective is unique in the specific field of unsupervised few-shot learning, which could not be directly borrowed to other fields.
\section{Our Method}

\subsection{Problem Formulation}
\label{problem_formulation}
The goal of \textit{unsupervised few-shot learning} is to first train a model on a large-scale \textit{unlabeled} auxiliary set $D_\text{train}$, and then apply this trained model on a novel labeled test set $D_\text{test}$, which is composed of a set of few-shot tasks. Note that, according to the setting of FSL, there are only a few labeled examples (\eg~$1$ or $5$ examples) in each class for each few-shot task in $D_\text{test}$. To effectively leverage the unlabeled auxiliary set $D_\text{train}$ for model training, following the episodic-training mechanism~\cite{Vinyals2016NIPS}, we still try to generate a series of pretext $N$-way $K$-shot tasks (episodes) from $D_\text{train}$ by using an data augmentation framework. In particular, each pretext few-shot task is composed of a pseudo support set (for training) and a pseudo query set (for validation). The pseudo support set consists of $N$ classes and $K$ examples per class (\eg~$K$=1 in our paper), termed as $\{(x_{i},y_{i})\}^{N \times K}_{i=1}$, while the query set $\{(\hat{x}_1, \hat{y}_2), ..., (\hat{x}_M, \hat{y}_M)\}$ contains $M$ generated examples augmented based on the pseudo support set. At each iteration, the model is trained by one episode (task) to minimize the classification loss on query set according to support set. After tens of thousands of episodes training, the model is expected to reach convergence and perform well on novel few-shot tasks.

\label{method:ULDA}
\subsection{The Proposed ULDA Framework}
To detail the proposed \textit{Distribution Shift-based Data Augmentation (ULDA)} framework (see Figure~\ref{fig:framework}), we first layout the pretext few-shot task construction procedure in unsupervised few-shot learning. Next, we detail the two key modules in ULDA: (1) \emph{distribution shift-based data augmentation module}, (2) \emph{metric-based few-shot learning module}. 

\begin{figure*}[!tbp]
\vspace{-0.3cm}
\centering
\includegraphics[width=0.9\textwidth]{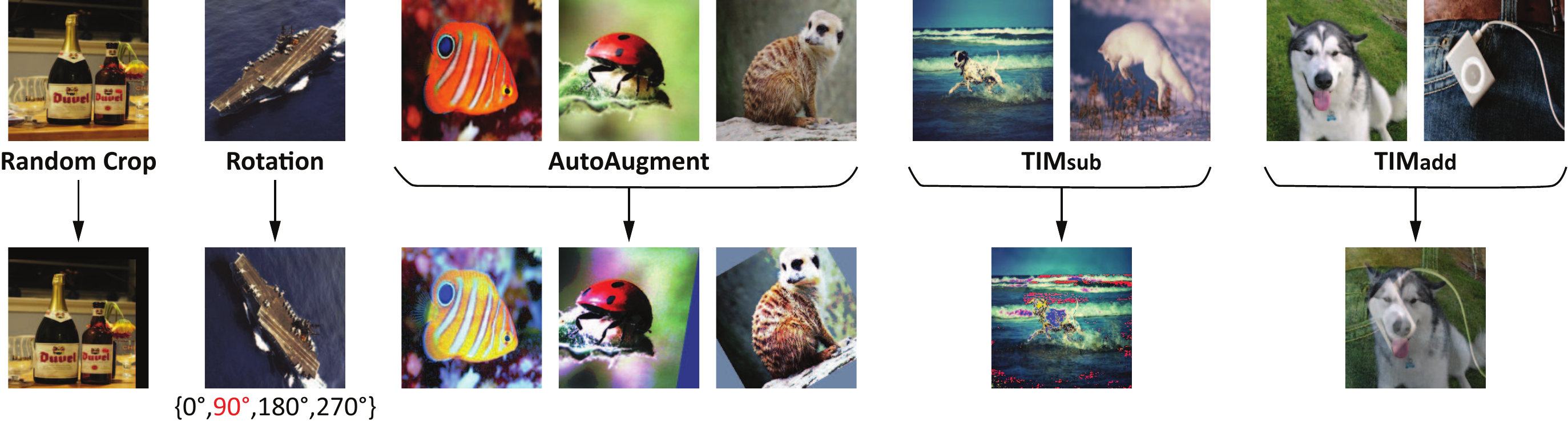}
\vspace{-0.2cm}
\caption{Illustrators of the employed augmentation techniques in this work. Top: Original images, Bottom: augmented images, transformed by an augmentation operator.}
\label{fig:transforms}
\vspace{-0.2cm}
\end{figure*}
 
\noindent\textbf{Task Construction for Unsupervised FSL.}
We randomly sample a mini-batch of $N$ data-points $\{x_1,...,x_N\}$ from the unlabeled auxiliary set $D_\text{train}$ as the initial support samples and construct one pretext few-shot task on augmented examples derived from this initial support set. Specifically, we take each data-point as one class and assign random labels for these data-points $X=\{(x_1,1),...,(x_N,N)\}$, which is a common strategy in unsupervised learning in the literature~\cite{Wu2018CVPR,He2019MoCo}.

During the augmentation on the support set, for the $i$-th initial support image $x_i$ (\ie~the $i$-th support class), we perform the augmentation operator $A^S_i$ from $\mathbb{A}^S$ ($A^S_i \in \mathbb{A}^S$) on this sample to obtain an augmented support image $A^S_i(x_i)$. Also, for the augmentation on the query set, we randomly select $M$ augmentation operators $A^Q_1$, $A^Q_2$, ..., $A^Q_M\in \mathbb{A}^Q$ to augment each initial support image (\ie~each support class) to obtain $M$ augmented query images. So each constructed pretext few-shot task $\mathcal{T}_z$ consists of an augmented support set $\mathcal{S}$ and an augmented query set $\mathcal{Q}$ (\ie $\mathcal{T}_z = (\mathcal{S}, \mathcal{Q})$):
\begin{equation}\small
\begin{split}
\mathcal{S} &= \big\{(A^S_i(x_i),i)|i=1,...,N\big\}, \\
\mathcal{Q} &= \big\{(A^Q_j(x_i),i)|i=1,...,N\, , \, j=1,...,M\big\},
\end{split}
\label{eqn:sampleSQ}
\end{equation}
where $A^S_i(x_i)$ means to perform the sampled operator $A^S_i$ on the $i$-th initial support image $x_i$ in the initial support set. $A^Q_j(x_i)$ means to perform the sampled operator $A^Q_j$ on the $i$-th initial support image $x_i$ from the initial support set.

In this work, we emphasize that maintaining a diversity between $\mathbb{A}^S$ and $\mathbb{A}^Q$ (\ie~$\mathbb{A}^S-\mathbb{A}^Q\neq\emptyset$ and $\mathbb{A}^Q-\mathbb{A}^S\neq\emptyset$) benefits the performance. This will be thoroughly discussed in the following section. We summarize the main sampling strategy of ULDA in Algorithm \ref{algorithm}.

\noindent\textbf{Distribution Shift-based Augmentation Module.}
Data augmentation technique plays a key role in aforementioned task construction procedure. However, in traditional methods~\cite{UMTRA2019NIPS, AAL2019ICML}, the generated tasks do not contain sufficient regularity for model learning as the generated examples are particularly suspect to visual similarity with the original images.
To alleviate this problem, we propose to increase the distribution diversity between the augmented support set and query set with a novel distribution shift-based data augmentation module, which employs diverse data augmentation operators to generate the support set and query set.


To systematically study the impact of diverse data augmentation, we consider to use both the commonly-used data augmentations and recently proposed augmentations.
Random crop and color jittering are widely used together in few-shot learning, we bind them as traditional augmentation (\textbf{TA} for short). Typically, for rotation, each image is converted among four directions in $\mathcal{R}=\{0^\circ,90^\circ,180^\circ,270^\circ\}$. The learned AutoAugment (\textbf{AA} for short) method proposed in~\cite{Cubuk2019CVPR} is also investigated for its promising performance in UMTRA~\cite{UMTRA2019NIPS}. 

In addition to the above existing augmentation techniques, we also propose a new \emph{Distribution Shift-based Task Internal Mixing (DSTIM)} augmentation strategy, which is composed of two  new augmentation operators, \ie~$\rm{TIM}_\text{sub}$ and $\rm{TIM}_\text{add}$. Specifically, we visualize all the augmentation operators used in this work in Figure~\ref{fig:transforms}. To understand the efficacy of each individual augmentation operator and the difference of different combinations of augmentation operators, we conduct  a serial of experiments detailed in our supplementary material.

\begin{table*}[!tp]\small
\centering
\extrarowheight=-1pt
\caption{Unsupervised few-shot classification results (\%) under $N$-way $K$-shot (\ie~(N, K)) setting on Omniglot. }
\label{tab:Omniglot}
\begin{tabular}{p{185pt}<{\raggedright}p{50pt}<{\raggedright}p{45pt}<{\centering}p{45pt}<{\centering}p{45pt}<{\centering}p{45pt}<{\centering}}
\toprule[1pt]
\textbf{Algorithms} &  \textbf{Clustering} & \textbf{(5, 1)} & \textbf{(5, 5)} &  \textbf{(20, 1)} & \textbf{(20, 5)} \\
\hline
\textbf{Training from scratch}                    & N/A           & 52.50\scalebox{0.75}{$\pm0.84$}  & 74.78\scalebox{0.75}{$\pm0.69$}  & 24.91\scalebox{0.75}{$\pm0.33$}  & 47.62\scalebox{0.75}{$\pm0.44$} \\
\hline
\small{\textbf{$\bm{k_{nn}}$-nearest neighbors}}  & DeepCluster   & 49.55\scalebox{0.75}{$\pm1.27$}  & 68.06\scalebox{0.75}{$\pm0.71$}  & 27.37\scalebox{0.75}{$\pm0.33$}  & 46.70\scalebox{0.75}{$\pm0.36$} \\
\textbf{linear classifier}                        & DeepCluster   & 48.28\scalebox{0.75}{$\pm1.25$}  & 68.72\scalebox{0.75}{$\pm0.66$}  & 27.80\scalebox{0.75}{$\pm0.61$}  & 45.82\scalebox{0.75}{$\pm0.37$} \\
\textbf{MLP with dropout}                         & DeepCluster   & 40.54\scalebox{0.75}{$\pm0.79$}  & 62.56\scalebox{0.75}{$\pm0.79$}  & 19.92\scalebox{0.75}{$\pm0.32$}  & 40.71\scalebox{0.75}{$\pm0.40$} \\
\textbf{cluster matching}                         & DeepCluster   & 43.96\scalebox{0.75}{$\pm0.80$}  & 58.62\scalebox{0.75}{$\pm0.78$}  & 21.54\scalebox{0.75}{$\pm0.32$}  & 31.06\scalebox{0.75}{$\pm0.37$} \\
\textbf{AAL-ProtoNes}~\cite{AAL2019ICML}          & N/A           & 84.66\scalebox{0.75}{$\pm0.70$}  & 88.41\scalebox{0.75}{$\pm0.27$}  & 68.79\scalebox{0.75}{$\pm1.03$}  & 74.05\scalebox{0.75}{$\pm0.46$} \\
\textbf{AAL-MAML++}~\cite{AAL2019ICML}            & N/A           & 88.40\scalebox{0.75}{$\pm0.75$}  & \textbf{97.96}\scalebox{0.75}{$\bm{\pm0.32}$}  & 70.21\scalebox{0.75}{$\pm0.27$}  & 88.32\scalebox{0.75}{$\pm1.22$} \\
\textbf{CACTUs-ProtoNets}~\cite{Hsu2019ICLR}      & ACAI          & 68.12\scalebox{0.75}{$\pm0.84$}  & 83.58\scalebox{0.75}{$\pm0.61$}  & 47.75\scalebox{0.75}{$\pm0.43$}  & 66.27\scalebox{0.75}{$\pm0.37$} \\
\textbf{CACTUs-MAML}~\cite{Hsu2019ICLR}           & ACAI          & 68.84\scalebox{0.75}{$\pm0.80$}  & 87.78\scalebox{0.75}{$\pm0.50$}  & 48.09\scalebox{0.75}{$\pm0.41$}  & 73.36\scalebox{0.75}{$\pm0.34$} \\
\textbf{UMTRA}~\cite{UMTRA2019NIPS}               & N/A           & 83.80\scalebox{0.75}{$\pm -  $}  & 95.43\scalebox{0.75}{$\pm -  $}  & 74.25\scalebox{0.75}{$\pm -  $}  & 92.12\scalebox{0.75}{$\pm -  $} \\
\hline
\textbf{ULDA-ProtoNets}(\textbf{ours})               & N/A           & \textbf{91.00}\scalebox{0.75}{$\bm{\pm0.42}$}  & \textbf{98.14}\scalebox{0.75}{$\bm{\pm0.15}$}  & \textbf{78.05}\scalebox{0.75}{$\bm{\pm0.31}$}  & \textbf{94.08}\scalebox{0.75}{$\bm{\pm0.13}$} \\
\textbf{ULDA-MetaOptNet}(\textbf{ours})            & N/A           & \textbf{90.51}\scalebox{0.75}{$\bm{\pm0.45}$}  & 97.60\scalebox{0.75}{$\pm0.17$}  & \textbf{76.32}\scalebox{0.75}{$\bm{\pm0.32}$}  & \textbf{92.48}\scalebox{0.75}{$\bm{\pm0.15}$} \\
\hline
\multicolumn{6}{c}  {\textit{Supervised (Upper Bound)}} \\
\hline
ProtoNets               & N/A           & 98.35\scalebox{0.75}{$\pm0.22$} & 99.58\scalebox{0.75}{$\pm0.09$}   & 95.31\scalebox{0.75}{$\pm0.18$}  & 98.81\scalebox{0.75}{$\pm0.07$} \\
MAML                    & N/A           & 94.46\scalebox{0.75}{$\pm0.77$} & 98.83\scalebox{0.75}{$\pm0.12$}   & 84.60\scalebox{0.75}{$\pm0.32$}  & 96.29\scalebox{0.75}{$\pm0.13$} \\
\bottomrule[1pt]
\end{tabular}
\end{table*}

\begin{algorithm}[t]
\caption{The main sampling strategy in ULDA}
\label{algorithm}
\hspace*{0.02in}{\bf require:}
$N$: class-count, $M$: meta-test size, $Z$: episodic number\\
\hspace*{0.02in}{\bf require:}
$\mathcal{U}$: unlabeled auxiliary set\\
\hspace*{0.02in}{\bf require:}
$\mathbb{A}^S, \mathbb{A}^Q$: two sets of different augmentation operators
\begin{algorithmic}[1]
\For{$z$ = $1,...,Z$}
\State Sample $N$ data-points $x_1,...,x_N$ from $\mathcal{U}$.
\State Randomly assign labels to sampled data-points: $X$=$\{(x_1,1),...,(x_N,N)\}$.
\State Generate support set $\mathcal{S}$ by using operator sampled from $\mathbb{A}^S$ to augment each sample in $X$.
\State Generate query set $\mathcal{Q}$ by using $M$ operator sampled from $\mathbb{A}^Q$ to augment each sample in $X$.
\State $\mathcal{T}_z \leftarrow (\mathcal{S},\mathcal{Q})$
\EndFor
\State \Return $\mathcal{T}_z|_{z=1}^{Z} $
\end{algorithmic}
\end{algorithm}

\noindent\textbf{DSTIM.}
Inspired by the recent works of generating new examples near the boundary of a classifier in \cite{Zhang2018ICLRmixup,Qiao2019ICCV}, we originally propose a task-level augmentation technique which is termed as \emph{Distribution Shift-based Task Internal Mixing (DSTIM)}. DSTIM is a simple yet effective method consisting of two augmentation operators $\rm{TIM}_\text{sub}$ and $\rm{TIM}_\text{add}$, both of which perform convex combination differently between all images in the operated data set. To be specific, for each instance $(x_i,y_i)$ in support (or query) set, we randomly select another instance $(x_j,y_j)$ from the same set. $\rm{TIM}_\text{add}$ synthesizes a new example $(\tilde{x}_\text{add},\tilde{y})$ as follows:
\begin{equation}\small
\label{eqn:tim_add}
\tilde{x}_\text{add} = \lambda \cdot x_i + (1-\lambda)\cdot x_j,\quad \tilde{y}=y_i,
\end{equation}
where $\lambda=\max(\lambda,1-\lambda), \lambda \sim \text{Beta} (\alpha,\alpha)$, so $\lambda \in [0.5, 1.0]$. This means $\rm{TIM}_\text{add}$ can extends the distribution of the synthesized example to the margin of the selected two examples. In contrast, for $\rm{TIM}_\text{sub}$, the synthesized example $\tilde{x}_\text{sub}$ can be obtained as below:
\begin{equation}\small
\label{eqn:tim_sub}
\tilde{x}_\text{sub} = \lambda \cdot x_i - (1.5-\lambda)\cdot x_j,\quad \tilde{y}=y_i,
\end{equation}
where $\lambda=0.5+\max(\lambda,1-\lambda), \lambda \sim \text{Beta} (\alpha,\alpha)$, so $\lambda \in [1.0, 1.5]$. 
$\rm{TIM}_\text{sub}$ can generate a new instance by performing subtraction between two images. And this operation can extend to get away from other examples.

Combining with these two operators, we can extend the distribution of raw examples to two opposite directions which thus strengthen the distribution shift between the two operated sets. Moreover, as we keep the value of $\lambda$ between 0.5 and 1.5, this will leads to the synthetic label $y_i$ rather than $y_j$, so it is an identity-preserved augmentation. In this work, we use $\rm{TIM}_\text{sub}$ to augment images in the support set and $\rm{TIM}_\text{add}$ for the query set.

\noindent\textbf{Metric-based FSL Module.}
Metric-based few-shot learning algorithms are a kind of simple and effective methods to address the few-shot problems, which aim to enhance the discriminability of learned feature representations via deep metric learning. The main component of these algorithms is a feature extractor $f(\cdot;\theta)$, which is a convolutional neural network (CNN) with parameters $\theta$. Given an episode (few-shot task) $\mathcal{T}_z$, the feature extractor will map each image $x_i$ in $\mathcal{T}_z$ into a $d$-dimensional feature, \textit{i.e.,} $f(x_i;\theta)$. In the learned feature space, the images in query set are forced to a labeled image in support set when they share similar semantic information~\cite{Sung2018CVPR,Li2019DN4}. Normally, Euclidean distance or cosine distance is employed to measure the distance or similarity between two examples. As the feature extractor plays a key role in the final classification results, the diversity of the augmented examples is crucial to exhibit the feature extractor to extract discriminative features. Crucially, our proposed ULDA framework can just satisfy this purpose, by increasing the distribution diversity between the augmented support set and query set. Therefore, to construct an effective unsupervised few-shot learning model, we tailor our ULDA into a representative existing metric-based few-shot learning algorithm, ProtoNets~\cite{Snell2016NIPS}, and name this new model as ULDA-ProtoNets. Obviously, our ULDA framework is universal and extensible, which can be simply tailored to other existing few-shot models. This part will be further discussed in Section~\ref{sec:extension}.

Given a \emph{N}-way \emph{K}-shot episode $\mathcal{T}_z$, ProtoNets computes the ``prototype'' via averaging features for each class in the support set with the feature extractor $f(\cdot;\theta)$:
\begin{equation}\small
\label{eqn:proto_type}
\mathbf{p}_i = \frac{1}{K}\sum_{x\in \mathcal{S}^i} f\big(A^S(x);\theta\big)\,,
\end{equation}
where $\mathcal{S}^i=\{x|(x,y)\in \mathcal{S},y=i\}$ and $A^S \in \mathbb{A}^S$. These ``prototypes'' are used to build a simple nearest neighbor classifier. Then, given a new image $x_q$ from query set, the classifier outputs a normalized classification score computed with Euclidean distance for each class $i$:
\begin{equation}\small
\label{eqn:proto_classification}
C^i\big(f(x_q;\theta)\big) = \frac{\Big\|f\big(A^Q(x_q);\theta\big) - \mathbf{p}_i\Big\|^2}{\sum_{j=1}^N \Big\|f\big(A^Q(x_q);\theta\big) - \mathbf{p}_j\Big\|^2}\,,
\end{equation}
where $A^Q \in \mathbb{A}^Q$. So, the image $x_q$ will be classified to its closest prototype. The few-shot loss function $\mathcal{L}_\text{few}$ for updating the parameter $\theta$ is formalized as:
\begin{equation}\small
\label{eqn:proto_loss_function}
\mathcal{L}_\text{few} = \sum_{\mathcal{T}_z\sim\mathcal{T}}\sum_{(x_q,y_q\in \mathcal{Q})} -\log C^{y_q}\big(f(x_q;\theta)\big)\,.
\end{equation}

Note that, the distance between $f(A^Q(x_q);\theta)$ and its corresponding prototype will not change if we keep $\mathbb{A}^S=\mathbb{A}^Q$. And this makes no sense to secure the discriminability of the feature extractor. Besides, as we use rotation as an augmentation technique, we can also incorporate with a self-supervised loss $\mathcal{L}_\text{self}$ to predict the rotation angle. 
\begin{equation}\small
\label{eqn:self_loss_function}
\mathcal{L}_\text{self} = \sum_{\mathcal{T}_z\sim\mathcal{T}}\sum_{(x_q,y_q\in \mathcal{Q})}^{y_\text{self}\in \mathcal{R}} -\!\log C^{y_\text{self}} (f(x_q;\theta,W)\,,
\end{equation}
where $W$ is the parameters of an additional classifier for predicting the rotation angle for each query image $x_q$ from $\mathcal{R}=\{0^\circ,90^\circ,180^\circ,270^\circ\}$. Specifically, this classifier is implemented by one fully connected layer.

Therefore, the overall loss function can be formulated as:
\begin{equation}\small
\label{eqn:total_loss_function}
\mathcal{L} = \mathcal{L}_\text{few} + \gamma \mathcal{L}_\text{self},
\end{equation}
where $\gamma$ is a balancing hyper-parameter.

\begin{table*}[!tp]\small
\centering
\extrarowheight=-1pt
\caption{Unsupervised few-shot classification results (\%) under $N$-way $K$-shot (\ie~(N,~K)) setting on \emph{mini}ImageNet. ``-'' means the results are not reported in their source papers.}
\label{tab:miniImageNet}
\vspace{-0.2cm}
\begin{tabular}{p{185pt}<{\raggedright}p{50pt}<{\raggedright}p{45pt}<{\centering}p{45pt}<{\centering}p{45pt}<{\centering}p{45pt}<{\centering}}
\toprule[1pt]
\textbf{Algorithms} &  \textbf{Clustering} & \textbf{(5, 1)} & \textbf{(5, 5)} &  \textbf{(5, 20)} & \textbf{(5, 50)} \\
\hline
\textbf{Training from scratch}                    & N/A           & 27.59\scalebox{0.75}{$\pm0.59$}  & 38.48\scalebox{0.75}{$\pm0.66$}  & 51.53\scalebox{0.75}{$\pm0.72$}  & 59.63\scalebox{0.75}{$\pm0.74$} \\
\hline
\small{\textbf{$\bm{k_{nn}}$-nearest neighbors}}  & DeepCluster   & 28.90\scalebox{0.75}{$\pm1.25$}  & 42.25\scalebox{0.75}{$\pm0.67$}  & 56.44\scalebox{0.75}{$\pm0.43$}  & 63.90\scalebox{0.75}{$\pm0.38$} \\
\textbf{linear classifier}                        & DeepCluster   & 29.44\scalebox{0.75}{$\pm1.22$}  & 39.79\scalebox{0.75}{$\pm0.64$}  & 56.19\scalebox{0.75}{$\pm0.43$}  & 65.28\scalebox{0.75}{$\pm0.34$} \\
\textbf{MLP with dropout}                         & DeepCluster   & 29.03\scalebox{0.75}{$\pm0.61$}  & 39.67\scalebox{0.75}{$\pm0.69$}  & 52.71\scalebox{0.75}{$\pm0.62$}  & 60.95\scalebox{0.75}{$\pm0.63$} \\
\textbf{cluster matching}                         & DeepCluster   & 22.20\scalebox{0.75}{$\pm0.50$}  & 23.50\scalebox{0.75}{$\pm0.52$}  & 24.97\scalebox{0.75}{$\pm0.54$}  & 26.87\scalebox{0.75}{$\pm0.55$} \\
\textbf{AAL-ProtoNes}~\cite{AAL2019ICML}          & N/A           & 37.67\scalebox{0.75}{$\pm0.39$}  & 40.29\scalebox{0.75}{$\pm0.68$}  & -      & - \\
\textbf{AAL-MAML++}~\cite{AAL2019ICML}            & N/A           & 34.57\scalebox{0.75}{$\pm0.74$}  & 49.18\scalebox{0.75}{$\pm0.47$}  & -      & - \\
\textbf{CACTUs-ProtoNets}~\cite{Hsu2019ICLR}      & DeepCluster   & 39.18\scalebox{0.75}{$\pm0.71$}  & 53.36\scalebox{0.75}{$\pm0.70$}  & 61.54\scalebox{0.75}{$\pm0.68$}  & 63.55\scalebox{0.75}{$\pm0.64$} \\
\textbf{CACTUs-MAML}~\cite{Hsu2019ICLR}           & DeepCluster   & 39.90\scalebox{0.75}{$\pm0.74$}  & 53.97\scalebox{0.75}{$\pm0.70$}  & \textbf{63.84}\scalebox{0.75}{$\bm{\pm0.70}$}  & \textbf{69.64}\scalebox{0.75}{$\bm{\pm0.63}$} \\
\textbf{UMTRA}~\cite{UMTRA2019NIPS}               & N/A           & 39.93\scalebox{0.75}{$\pm -  $}  & 50.73\scalebox{0.75}{$\pm -  $}  & 61.11\scalebox{0.75}{$\pm -  $}  & 67.15\scalebox{0.75}{$\pm -  $} \\
\hline
\textbf{ULDA-ProtoNets}(\textbf{ours})              & N/A           & \textbf{40.63}\scalebox{0.75}{$\bm{\pm0.61}$}  & \textbf{56.18}\scalebox{0.75}{$\bm{\pm0.59}$}  & \textbf{64.31}\scalebox{0.75}{$\bm{\pm0.51}$}  & 66.43\scalebox{0.75}{$\pm0.47$} \\
\textbf{ULDA-MetaOptNet}(\textbf{ours})           & N/A           & \textbf{40.71}\scalebox{0.75}{$\bm{\pm0.62}$}  & \textbf{54.49}\scalebox{0.75}{$\bm{\pm0.58}$}  & 63.58\scalebox{0.67}{$\pm0.51$}  & \textbf{67.65}\scalebox{0.67}{$\bm{\pm0.48}$} \\
\hline
\multicolumn{6}{c}  {\textit{Supervised (Upper Bound)}} \\
\hline
ProtoNets               & N/A           & 46.56\scalebox{0.75}{$\pm0.76$} & 62.29\scalebox{0.75}{$\pm0.71$}   & 70.05\scalebox{0.75}{$\pm0.65$}  & 72.04\scalebox{0.75}{$\pm0.60$} \\
MAML                    & N/A           & 46.81\scalebox{0.75}{$\pm0.77$} & 62.13\scalebox{0.75}{$\pm0.72$}   & 71.03\scalebox{0.75}{$\pm0.69$}  & 75.54\scalebox{0.75}{$\pm0.62$} \\
\bottomrule[1pt]
\end{tabular}
\end{table*}

\begin{table*}[!tbp]\small
\begin{center}
\caption{Unsupervised few-shot classification results in \% of $N$-way $K$-shot (N,~K) learning methods on \emph{tiered}ImageNet.}
\vspace{-0.2cm}
\label{tab:tieredImageNet}
\small{
\begin{tabular}{p{185pt}<{\raggedright}p{50pt}<{\raggedright}p{45pt}<{\centering}p{45pt}<{\centering}p{45pt}<{\centering}p{45pt}<{\centering}}
\toprule[1pt]
\textbf{Algorithms}       & \textbf{Clustering}  & \textbf{(5, 1)} & \textbf{(5, 5)} &  \textbf{(5, 20)} & \textbf{(5, 50)} \\
\hline
{\textbf{Training from scratch}}  & N/A  & 26.27\scalebox{0.75}{$\pm1.02$} & 34.91\scalebox{0.75}{$\pm0.63$} & 38.14\scalebox{0.75}{$\pm0.58$} & 38.67\scalebox{0.75}{$\pm0.44$} \\
\hline
{\textbf{ULDA-ProtoNets(ours)}}          & N/A  & \textbf{41.60}\scalebox{0.67}{$\bm{\pm0.64}$} & \textbf{56.28}\scalebox{0.67}{$\bm{\pm0.62}$} & \textbf{64.07}\scalebox{0.67}{$\bm{\pm0.55}$} & \textbf{66.00}\scalebox{0.67}{$\bm{\pm0.54}$} \\
{\textbf{ULDA-MetaOptNet(ours)}}         & N/A  & \textbf{41.77}\scalebox{0.67}{$\bm{\pm0.65}$} & \textbf{56.78}\scalebox{0.67}{$\bm{\pm0.63}$} & \textbf{67.21}\scalebox{0.67}{$\bm{\pm0.56}$} & \textbf{71.39}\scalebox{0.67}{$\bm{\pm0.53}$} \\
\hline
\multicolumn{6}{c}  {\emph{Supervised (Upper Bound)}} \\
\hline
ProtoNets         & N/A  & 46.66\scalebox{0.75}{$\pm0.63$} & 66.01\scalebox{0.75}{$\pm0.60$} & 77.62\scalebox{0.75}{$\pm0.46$} & 81.70\scalebox{0.75}{$\pm0.44$} \\
MetaOptNet        & N/A  & 47.32\scalebox{0.75}{$\pm0.64$} & 66.16\scalebox{0.75}{$\pm0.58$} & 77.68\scalebox{0.75}{$\pm0.47$} & 80.61\scalebox{0.75}{$\pm0.48$} \\
\bottomrule[1pt]
\end{tabular}
}
\end{center}
\end{table*}

\subsection{Extension to Optimization-based FSL}
\label{sec:extension}
Different from the metric-based FSL algorithms, the optimization-based FSL algorithms strive for enhancing the flexibility of a few-shot model such that it can be readily updated using a few labeled examples. Most of these algorithms are generally based on meta learning. See Section \ref{sec:related_work} for more details. To further verify the effectiveness and scalability of our proposed ULDA framework, we extend ULDA to a recently proposed optimization-based FSL algorithm, \ie~MetaOptNet~\cite{LeeCVPR2019meta}, and name this new method as ULDA-MetaOptNet in the following parts.


\section{Experiments}
\label{sec:experiments}

In this section, we detail the experimental settings and compare our ULDA with the state-of-the-art approaches on two challenging datasets, \ie~Omniglot~\cite{lake2011one} and \emph{mini}ImageNet~\cite{Vinyals2016NIPS}, which are widely used in the literature.

\subsection{Experimental Setting}
\noindent\textbf{Datasets.} 
The \textbf{Omniglot} dataset comprises 1623 characters from 50 different alphabets. Each character contains 20 instances written by different persons. We follow the experiment protocol described by~\cite{AAL2019ICML}: classes 1-1150, 1150-1200 and 1200-1623 are used for training, validation and test, respectively.

The \textbf{\emph{mini}ImageNet} is the most popular benchmark in the field of few-shot learning, which was introduced in~\cite{Vinyals2016NIPS}. It is composed of 100 classes selected from ImageNet~\cite{Alex2012NIPS}, and each class contains 600 images with the size of $84\times84$. We follow the data splits proposed by~\cite{RaviICLR2017}, which splits the total 100 classes into 64 classes for training, 16 classes for validation and 20 classes for test, respectively.

\noindent\textbf{Backbone network.}
We employ a four-layer convolutional neural network as the feature extractor backbone, which is widely adopted in the few-shot learning literature~\cite{Snell2016NIPS,FinnICML2017}. Each layer comprises a 64 filters ($3\times3$ kernel) convolutional layer, a batch normalization layer, a ReLU layer and a $2\times2$ max-pooling layer. Moreover, different ResNet \cite{He2016ResNet} architectures are also employed to validate the expansibility of our framework.

\begin{figure*}[!tbp]
\begin{center}
  \includegraphics[width=0.90\textwidth]{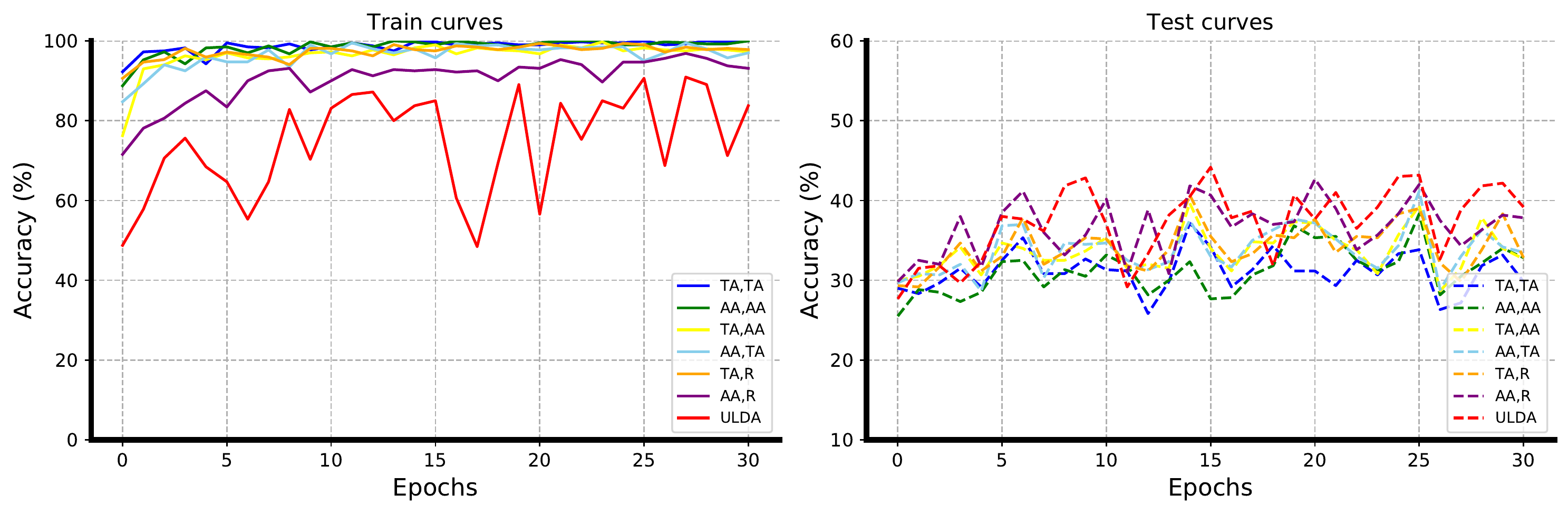}
\end{center}
\vspace{-0.3cm}
  \caption{The train and test accuracy curves on the 5-way 1-shot tasks. As seen, the diverse combinations especially our proposed ULDA (the red lines) enjoys a smaller risk of overfitting and a higher test accuracy.}
  \label{fig:total_train_test_curves}
\end{figure*}

\noindent\textbf{Training strategy.}
We conduct $N$-way $K$-shot classification tasks on the aforementioned datasets. We randomly sample and construct 10,000 pretext few-shot tasks in each epoch and train our networks for a total of 60 epochs. For \emph{mini}ImageNet, we employ AutoAugment~\cite{Cubuk2019CVPR} to augment the support set and traditional augmentation together with rotation to augment the query set. For Omniglot, we use AutoAugment for support set and random crop for query set. Note that, self-supervised loss is not employed in Omniglot. All backbone networks are optimized by Adam~\cite{Kingma2014Adam}. The initial learning rate is set as 0.001 and multiplied by 0.06, 0.012, 0.0024 after 20, 40, and 50 epochs, respectively. We conduct all the experiments on GTX 2080Ti. For a fair comparison, the hyper parameters in all of these methods are kept to be the same.

\noindent\textbf{Parameter setup.} We set $\gamma=1$ in Eq.~(\ref{eqn:total_loss_function}). In Eq.~(\ref{eqn:tim_add}), we empirically set $\alpha$=0.8 for $\rm{TIM}_\text{sub}$ and in Eq.~(\ref{eqn:tim_sub}), $\alpha$=0.6 for $\rm{TIM}_\text{add}$. Our model is robust to different values of $\alpha$ according to our experiments (see more details in our supplementary material). Thus, we set it in a slightly different manner following our distribution-diversity argument.

\subsection{Unsupervised Few-shot Learning Results}
To verify the effectiveness of our approach for unsupervised few-shot learning, we compare our framework with the state-of-the-art (SOTA) methods in various settings. Moreover, to make our results more convincing, we randomly sample 1,000 episodes from the test set for evaluation. Also, we take the top-1 mean accuracy as evaluation criterion and repeat this process five times. Besides, the $95\%$ confidence intervals are also reported.

\noindent\textbf{Results on Omniglot.}
The comparative results between a variety of baseline and recently proposed methods on Omniglot are presented in Table \ref{tab:Omniglot}. ULDA shows currently the best results across different tasks. Compared with previous best results, our ULDA-ProtoNets gains 2.6\%, 0.18\%, 3.8\% and 1.96\% under 5-way 1-shot, 5-way 5-shot, 20-way 1-shot and 20-way 5-shot settings, respectively. Similarly, our ULDA-MetaOptNet can also achieve very competitive results especially in the 5-way 1-shot and 20-way 1-shot settings.

\noindent\textbf{Results on \emph{mini}ImageNet.}
The experimental results on \emph{mini}ImageNet are summarized in Table \ref{tab:miniImageNet}. Our ULDA achieves the state-of-the-art results on both 5-way 1-shot, 5-way 5-shot and 5-way 20-shot settings and achieves competitive results on 5-way 50-shot settings. 
Besides, the results of ULDA are very close to the results of supervised few-shot learning approaches with a labeled auxiliary set, \ie~ProtoNets and MAML. Note that, when using the same few-shot learning algorithm (\ie~ProtoNets), our ULDA framework outperforms all other methods across different classification tasks. Compared with CACTUs-ProtoNets, our ULDA-ProtoNets gains 1.45\%, 2.82\%, 2.77\%, 2.88\% performance boost under 5-way 1-shot, 5-shot, 20-shot and 50-shot settings, respectively. The reason is that CACTUS uses clustering algorithms to obtain the pseudo labels before constructing few-shot tasks, but the quality of these pseudo labels will limit the final results. In contrast, our ULDA does not have this limitation. When compared with AAL, which is the closest work to ours, our ULDA can still achieve 2.96\% and 15.89\% performance boost for 5-way 1-shot and 5-way 5-shot, respectively.

\textbf{Results on \emph{tiered}ImageNet.}
We turn to \emph{tiered}ImageNet, a more challenging dataset, which contains more complex classes and examples than \emph{mini}ImageNet. Since the recent unsupervised few-shot leaning methods (\ie~CACTUs, UMTRA) did not report their results on this dataset, we only compare our methods with the baseline method \textit{training from scratch}. The results are illustrated in Table \ref{tab:tieredImageNet}. Our ULDA performs much better than learning from scratch and slightly weaker than the supervised methods.

\begin{figure}[!tbp]
\centering
\includegraphics[width=0.40\textwidth]{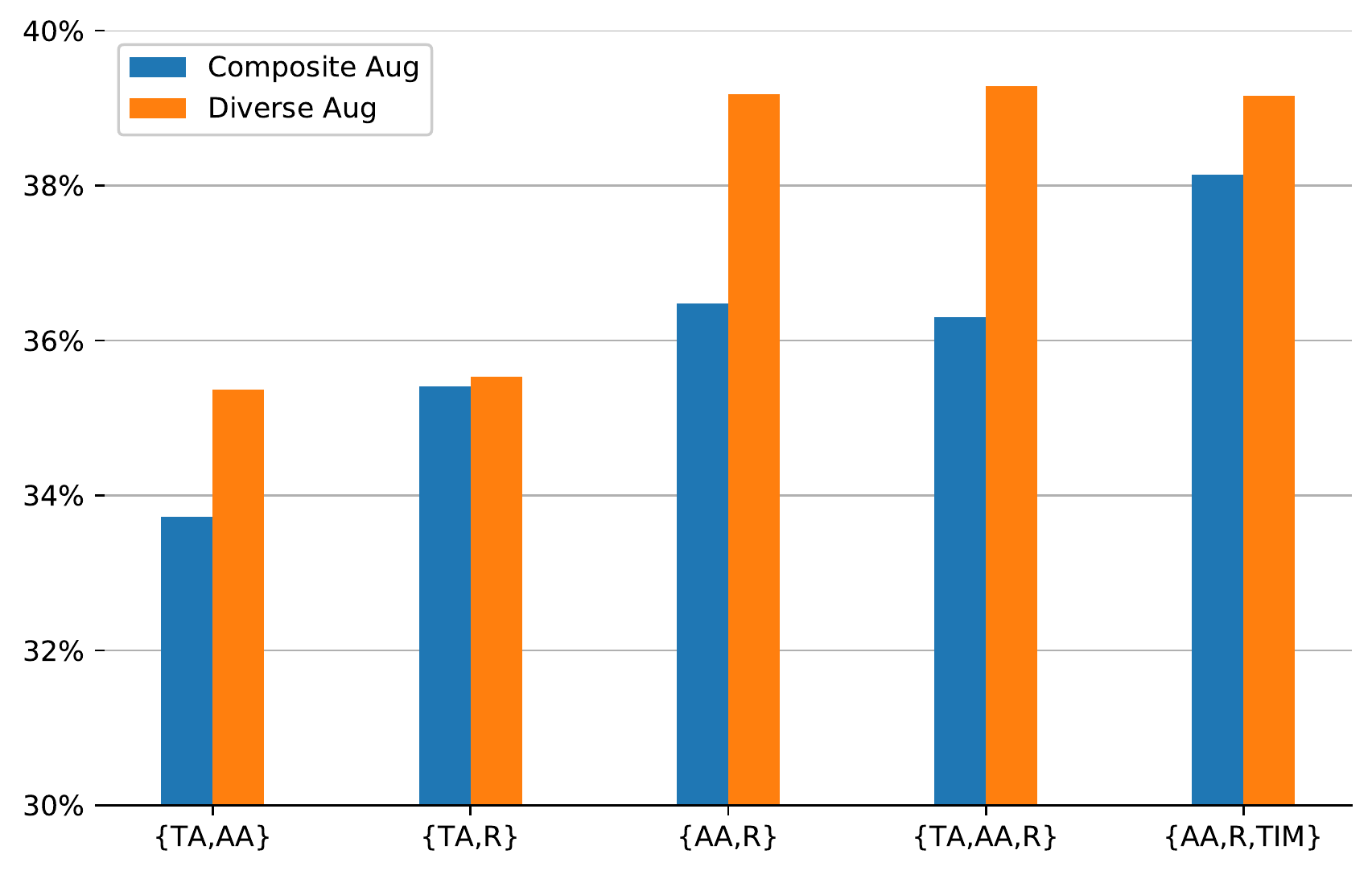}
\vspace{-0.3cm}
\caption{Comparison between composite augmentation and diverse augmentation on \emph{mini}ImageNet.}
\label{fig:com_sep}
\end{figure}

\begin{figure}[!tbp]
\centering
\includegraphics[width=0.40\textwidth]{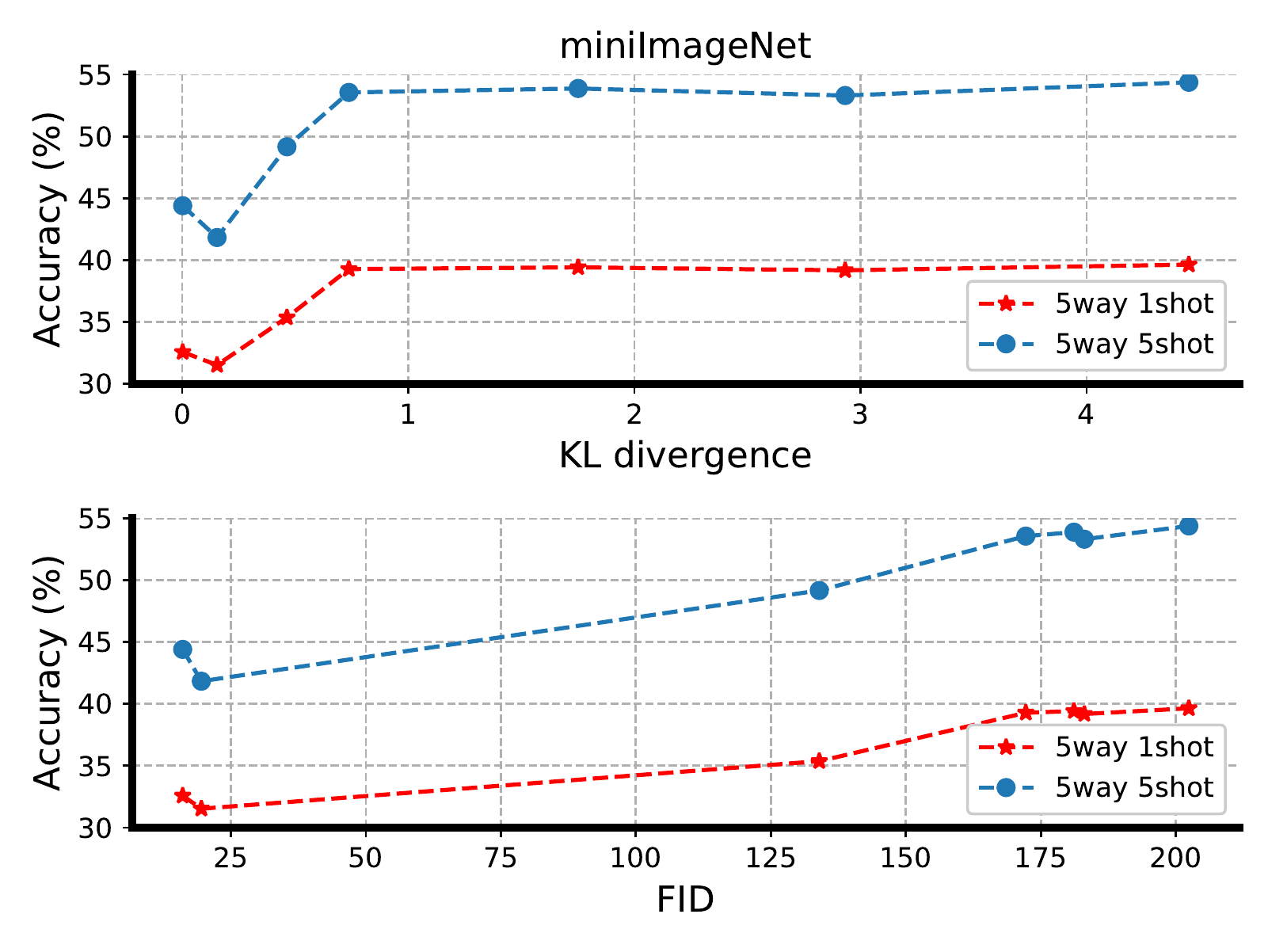}
\vspace{-0.3cm}
\caption{The performance changing with the value of distribution divergence on \emph{mini}ImageNet.}
\label{fig:distribution_difference}
\vspace{-0.2cm}
\end{figure}

\subsection{Ablation Study on \emph{mini}ImageNet}
\label{sec:ablation_study}

\noindent\textbf{The Overfitting Problem}
In these series of experiments, we study the overfitting problem of different diverse augmentation combinations during the model learning procedure. The results on \emph{mini}ImageNet under 5-way 1-shot are shown in Figure \ref{fig:total_train_test_curves}. Here, ULDA is the proposed framework in this paper, which employs AutoAugment to generate support set and combines traditional augmentation with rotation to generate query set. Moreover, DSTIM is also employed here. All results are averaged among 1,000 tasks. As expected, diverse augmentation can efficiently alleviate the over-fitting problem. Moreover, when incorporated with our proposed augmentation method DSTIM, the distribution difference between query set and support set can be further enlarged, \ie the generated pretext few-shot tasks enjoy more challenges, which can effectively alleviate the overfitting problem in unsupervised learning manner. As seen in Figure \ref{fig:total_train_test_curves}, our proposed ULDA obtains a lower train accuracy curves but meanwhile a relative higher test accuracy curves.

\noindent\textbf{Composite Augmentation vs. Diverse Augmentation.}
Another way to alleviate the overfitting problem in unsupervised FSL is that we can compose different augmentation operators together (\ie~a larger augmentation operator set $\mathbb{A}=\mathbb{A}^S \cup \mathbb{A}^Q$) to increase the whole diversity of the generated samples as introduced in \cite{Chen2020SimCLR}, but we still adopt the same $\mathbb{A}$ to augment both the query and support set. We call this \textit{composite augmentation}. Differently, our ULDA employs a \textit{diverse augmentation}, \ie~augmenting the query set and support set separately. To figure out the difference between these two augmentation ways, we conduct a serial of experiments on \emph{mini}ImageNet (see Figure \ref{fig:com_sep}). When we employ more complex operators, both the diverse augmentation and composite augmentation boost the performance. Notably, the diverse augmentation always performs better than the composite augmentation. It shows that the former can gain more distribution shift, which is more beneficial for alleviating the overfitting problem.


\begin{table}[tbp]\footnotesize
\begin{center}
\tabcolsep=2.3pt
\caption{5-way 1-shot accuracy (\%) on \emph{mini}ImageNet with different network architectures.}
\label{tab:different_architecture}
\vspace{-0.3cm}
\begin{tabular}{cccccc}
\toprule[1pt]
                & ResNet12  & ResNet18 & ResNet34 & ResNet50 \\
\hline
ULDA-ProtoNets  & \textbf{42.73}\scalebox{0.75}{$\bm{\pm0.62}$} & 42.05\scalebox{0.75}{$\pm0.56$} & 40.48\scalebox{0.75}{$\pm0.57$} & 39.48\scalebox{0.75}{$\pm0.56$} \\
\bottomrule[1pt]
\end{tabular}
\end{center}
\vspace{-0.3cm}
\end{table}

\noindent\textbf{Comparisons with different backbones.}
We further perform a series of experiments on ResNets~\cite{He2016ResNet} with different depths. Note that the settings are kept almost the same as the above experiments expect the learning rate. We set the learning rate to 0.1 following \cite{LeeCVPR2019meta}. The results are reported in Table~\ref{tab:different_architecture}. As seen, our ULDA can achieve much higher results with much deeper networks, \textit{e.g.,} ResNet12 and ResNet18. For example, when using ResNet12 as the backbone, our ULDA-ProtoNets can even further gain $2.1\%$ improvements over a Conv64F-based version, which is also significantly better than other SOTA methods.
However, the performance of our ULDA-ProtoNets begins to drop with ResNet18/ResNet34/ResNet50, which indicates that these models suffer from a new overfitting risk. We may need to further increase the diversity between the constructed tasks. We leave this as our future work.

\noindent\textbf{Effectiveness of distribution shift-based augmentation module.} Despite the promising results achieved by our entire framework, we also expect to know how it works, especially the relationship between the distribution shift in generated two sets and the final results.
With this purpose, we employ the aforementioned augmentation techniques (\ie~random crop, color jittering, rotation, AutoAugment and our proposed DSTIM) and combine them in various ways to produce these two sets with different distribution shift. Besides, we use Kullback-Leibler divergence (KL divergence) and Fr\'echet Inception Distance (FID)~\cite{Heusel2017NIPS} to evaluate the distribution difference. The results are illustrated in Figure \ref{fig:distribution_difference}.
We can draw the conclusion from these results that the models tend to perform much better when trained on pretext few-shot tasks that have large distribution difference. 

\begin{figure}[!tbp]
\centering
\includegraphics[width=0.42\textwidth]{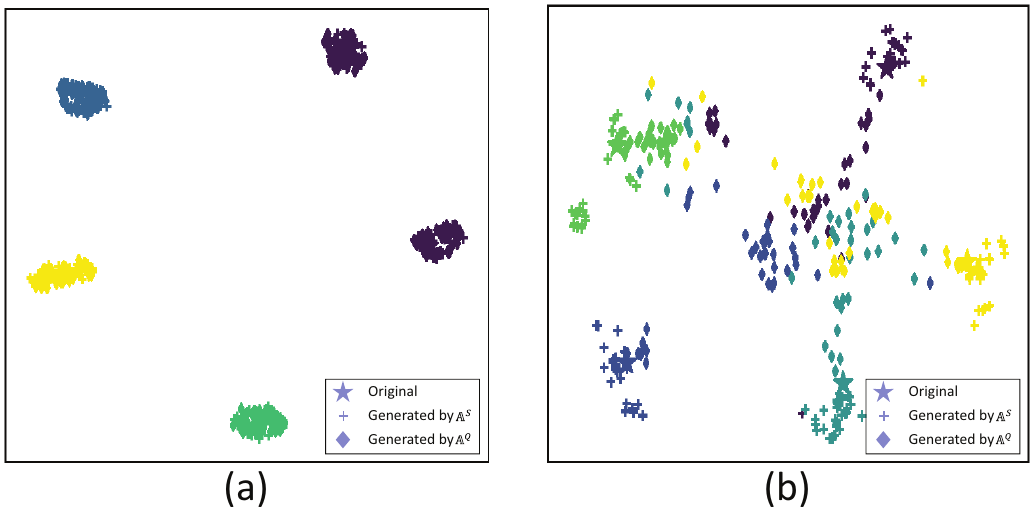}
\vspace{-0.3cm}
\caption{t-SNE plots in feature space. (a) common augmentation, (b) our ULDA. Zoom in for best visual effect.}
\label{fig:tsne_fea}
\vspace{-0.2cm}
\end{figure}

In order to intuitively show the effect of our framework, we also visualize the augmentation effect in feature space in Figure \ref{fig:tsne_fea}. As seen, when augmenting support set and query set with the same augmentation techniques, the generated query set gathers tightly around support set, and these tend to exist heavy overlap in these augmented data-points. However, with our ULDA, the generated examples share more diversity and more distribution difference between the support and query set. We will analyze this issue in supplementary material.

\section{Conclusion}
In this paper, we present an unsupervised few-shot learning framework that aims to increase the diversity of generated few-shot tasks based on data augmentation. We argue that when strengthening the distribution shift between the support set and query set in each few-shot task with different augmentation techniques can increase the generalization ability for model training. A serial of experiments have been conducted to demonstrate the correctness of our finding. We also incorporate our framework with two representative few-shot learning algorithms, \ie~ProtoNets and MetaOptNet, and achieve the state-of-the-art results across a variety of few-shot learning tasks established on Omniglot and \emph{mini}ImageNet.


\appendix   
\setcounter{table}{0}   
\setcounter{figure}{0}
\renewcommand{\thetable}{A\arabic{table}}
\renewcommand{\thefigure}{A\arabic{figure}}
\numberwithin{figure}{section}
\numberwithin{table}{section}

\section{The Comparison of Different Diverse Augmentation Combinations}
To verify the diverse augmentation which we claim on, we do a serial of experiments. Here, we employ the aforementioned augmentation techniques (\ie~TA, AA, R and our proposed DSTIM) and combine them in various ways to produce different distribution shift when constructing query and support set. Besides, we use Kullback-Leibler divergence (KL divergence) and Fr\'echet Inception Distance (FID) to measure the distribution difference. Also, the results in Table \ref{tab:difference_compare_augmentation} are the detailed values of Figure 5 in our original paper.
\begin{table}[htbp]\footnotesize
\caption{The comparison with different augmentation methods on \emph{mini}ImageNet. The results in \% of $N$-way $K$-shot ($N$, $K$) are reported.}
\label{tab:difference_compare_augmentation}
\begin{center}
\begin{tabular}{p{40pt}<{\raggedright}p{45pt}<{\raggedright}p{13pt}<{\centering}p{16pt}<{\centering}p{30pt}<{\centering}p{30pt}<{\centering}}
\toprule[1pt]
\textbf{$\mathbb{A}^S$} &  \textbf{$\mathbb{A}^Q$} & \textbf{KL} & \textbf{FID} &  \textbf{(5, 1)} & \textbf{(5, 5)} \\
\hline
TA & TA   & 0.00 & 16.07  & 32.58\scalebox{0.75}{$\pm 0.49$}  & 44.40\scalebox{0.75}{$\pm 0.49$} \\ 
AA & AA   & 0.15 & 19.52  & 31.53\scalebox{0.75}{$\pm 0.49$}  & 41.83\scalebox{0.75}{$\pm 0.53$} \\ 
TA & AA   & 0.41 & -      & 34.07\scalebox{0.75}{$\pm 0.51$}  & 47.31\scalebox{0.75}{$\pm 0.52$} \\ 
AA & TA   & 0.46 & 133.97 & 35.37\scalebox{0.75}{$\pm 0.53$}  & 49.16\scalebox{0.75}{$\pm 0.52$} \\ 
AA & R    & 0.73 & 183.06 & 39.18\scalebox{0.75}{$\pm 0.58$}  & 53.30\scalebox{0.75}{$\pm 0.58$} \\ 
AA & R+TA & 2.66 & 172.22 & 39.28\scalebox{0.75}{$\pm 0.59$}  & 53.55\scalebox{0.75}{$\pm 0.58$} \\ 
AA & R+$\rm{TIM}_{add}$    & 2.93 & 181.14 &  39.42\scalebox{0.75}{$\pm 0.57$} & 53.87\scalebox{0.75}{$\pm 0.58$} \\ 
AA+$\rm{TIM}_{sub}$ & R+$\rm{TIM}_{add}$     & 4.45   & 185.27  &39.52\scalebox{0.75}{$\pm 0.58$} & 54.26\scalebox{0.75}{$\pm 0.57$} \\ 
AA+$\rm{TIM}_{sub}$ & R+TA+$\rm{TIM}_{add}$  & \textbf{5.33} & \textbf{202.42} & \textbf{39.64}\scalebox{0.75}{$\bm{\pm 0.60}$}  & \textbf{54.37}\scalebox{0.75}{$\bm{\pm 0.58}$} \\ 
\bottomrule[1pt]
\end{tabular}
\end{center}
\vspace{-0.2cm}
\end{table}

\section{Feature Representation}

\begin{figure}[!tbp]
\begin{center}
  \includegraphics[width=0.48\textwidth]{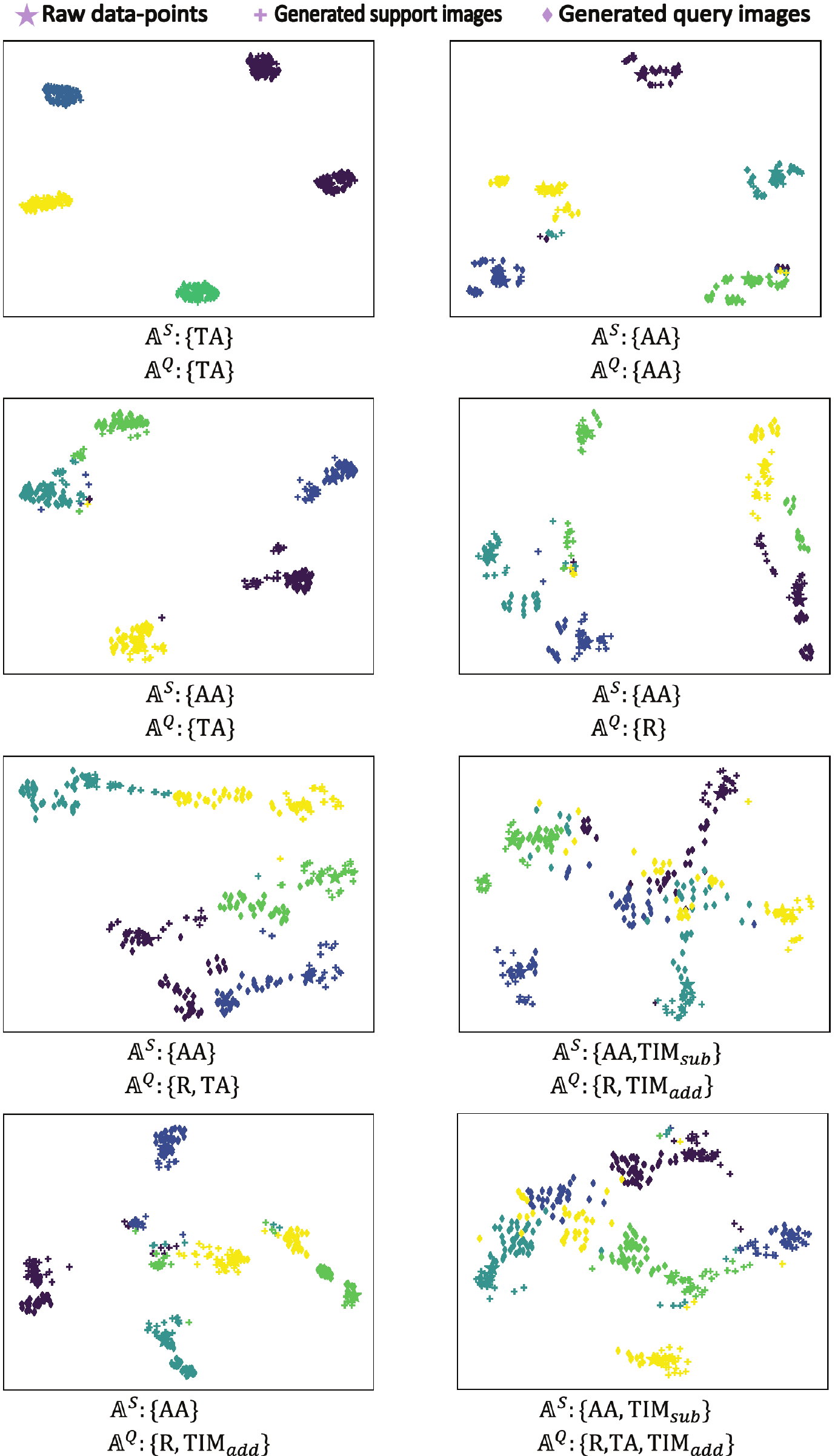}
\end{center}
  \caption{Visualization of feature transformations in generated support images and query images. Same color means generated from the same data-point. The generated images own more diversity and there exist little overlap between generated support images and query images via our approach.}
    \label{fig:tsne_feature}
\end{figure}

In order to intuitively show the effect of our framework, we also visualize the augmentation efficacy in feature space in Figure \ref{fig:tsne_feature}. We find that, when augmenting support set and query set using the same augmentation technique, the generated query set gathers tightly around support set, and this case tends to exist heavy overlap in these augmented data-points. However, by using our approach, the generated examples share more diversity and more distribution difference between the support set and query set.

\section{The Value of \texorpdfstring{$\alpha$}~~in DSTIM}
Our default setting is $\alpha=0.8$ for $\rm{TIM}_{sub}$ and $\alpha=0.6$ for $\rm{TIM}_{add}$. The performance remains stable with using different values of $\alpha$. The results are shown in Table \ref{tab:different_lambda}. 

\begin{table}[h]\small
\caption{The comparison with different augmentation methods on \emph{mini}ImageNet. The results in \% of $N$-way $K$-shot ($N$, $K$) are reported.}
\label{tab:different_lambda}
\vspace{-0.3cm}
\begin{center}
\begin{tabular}{cccc}
\toprule[1pt]
$\alpha$ for $\rm{TIM}_{sub}$ &  $\alpha$ for $\rm{TIM}_{add}$ & \textbf{(5, 1)} & \textbf{(5, 5)} \\
\hline
0.6 & 0.6 & 40.08 $\pm 0.59$ & 54.33 $\pm 0.56$ \\
0.6 & 0.8 & 40.06 $\pm 0.61$ & 54.47 $\pm 0.58$ \\
0.8 & 0.6 & \textbf{40.63} ${\bm{\pm 0.61}}$ & \textbf{55.41} ${\bm{\pm 0.57}}$ \\
0.8 & 0.8 & 39.92 $\pm 0.60$ & 54.76 $\pm 0.56$ \\
\bottomrule[1pt]
\end{tabular}
\end{center}
\vspace{-0.25cm}
\end{table}

\newpage
\bibliography{egbib}



\end{document}